\documentclass[letterpaper]{article} 
\usepackage[preprint]{aaai2027}
\usepackage[hyphens]{url}  
\usepackage{graphicx} 
\def\UrlFont{\rm}  
\usepackage{natbib}  
\usepackage{caption} 
\usepackage{algorithm}
\usepackage{algorithmic}
\usepackage{graphicx}
\usepackage{subcaption}

\usepackage{comment}

\usepackage{newfloat}
\usepackage{listings}
\usepackage{booktabs} 
\usepackage{multirow} 
\DeclareCaptionStyle{ruled}{labelfont=normalfont,labelsep=colon,strut=off} 
\floatstyle{ruled}
\newfloat{listing}{tb}{lst}{}
\floatname{listing}{Listing}

\usepackage{booktabs}

\usepackage{amsmath}
\usepackage{amssymb}
\usepackage{bm}

\usepackage[most]{tcolorbox}
\usepackage{xcolor}

\usepackage{cleveref}
\title{Cross-Branch Conflict as a Shield: Safeguarding Facial Identities \\ in Unified Multimodal Image Editing}
\author{
    Weiwei Tan\textsuperscript{\rm 1}\equalcontrib,
    Junxian Li\textsuperscript{\rm 2}\equalcontrib,
    Rui Wang\textsuperscript{\rm 1}\corresponding,
    Zhenhua Xu\textsuperscript{\rm 3},
    Yanjun Zhang\textsuperscript{\rm 4}, 
    Yu Leo Zhang\textsuperscript{\rm 4}
}
\affiliations{
    \textsuperscript{\rm 1} Institute of Information Engineering, Chinese Academy of Sciences \\

    \textsuperscript{\rm 2} Shanghai Jiao Tong University \\
    \textsuperscript{\rm 3} Zhejiang University \\
    \textsuperscript{\rm 4} Griffith University \\
    \texttt{\{tanweiwei, RuiWang\}@iie.ac.cn, lijunxian0531@sjtu.edu.cn, xuzhenhua0326@zju.edu.cn, \{yanjun.zhang, leo.zhang\}@griffith.edu.au}
}

\begin{document}

\maketitle

\begin{abstract}
Unified multimodal models (UMMs) have recently demonstrated powerful instruction-based image editing capabilities, while also raising serious concerns about the unauthorized manipulation of personal portraits. We investigate a novel and practical problem: protecting facial identities against unauthorized editing of UMMs. Existing diffusion-based and VLM-based protection methods often become ineffective because they typically disrupt only a single visual branch. To understand this limitation, we conduct a feature-level analysis of the understanding and generation branches in unified image editing models. Our observations show that the structural agreement between these two branches is closely related to successful image editing. When only one branch is distorted, the model may still recover identity information from the other branch. Based on this, we propose Cross-Branch Conflict as a Shield (CCS), a unified adversarial protection framework. CCS jointly drives the ViT and VAE representations away from their clean counterparts. It also uses a linear Centered Kernel Alignment (CKA) objective to disrupt the structural consistency between the two branches. By degrading reliable identity information in both visual pathways and inducing incompatible cross-branch representations, CCS effectively prevents UMMs from recovering consistent facial identity cues during editing. Extensive experiments suggest that CCS consistently provides stronger protection in suppressing identity-preserving edits. Codes are in the supplementary material.

\end{abstract}

\section{Introduction}
\label{sec:intro}

Unified multimodal models (UMMs)~\cite{deng2025emerging,chen2025janus,xie2026show,tian2026internvl,diao2026sensenova} are reshaping the landscape of visual modeling. Conventional vision-language models~\cite{Liu2023VisualIT, Zhu2023MiniGPT4EV,wang2024qwen2} (VLMs) mainly focus on visual understanding, while diffusion models~\cite{Ho2020DenoisingDP,Rombach2021HighResolutionIS} are mainly designed for image generation. In contrast, UMMs combine visual understanding, language reasoning, and image generation within one model. This design enables UMMs to produce realistic, identity-preserving facial edits from a portrait and a natural-language instruction. While useful for image creation, this nature also lowers the cost of facial-image misuse. Public portraits can be misused to generate false or unauthorized facial content. In response to this security issue, protecting facial images from unauthorized use is essential.

\begin{figure}[t]
    \centering
    \includegraphics[width=\columnwidth]{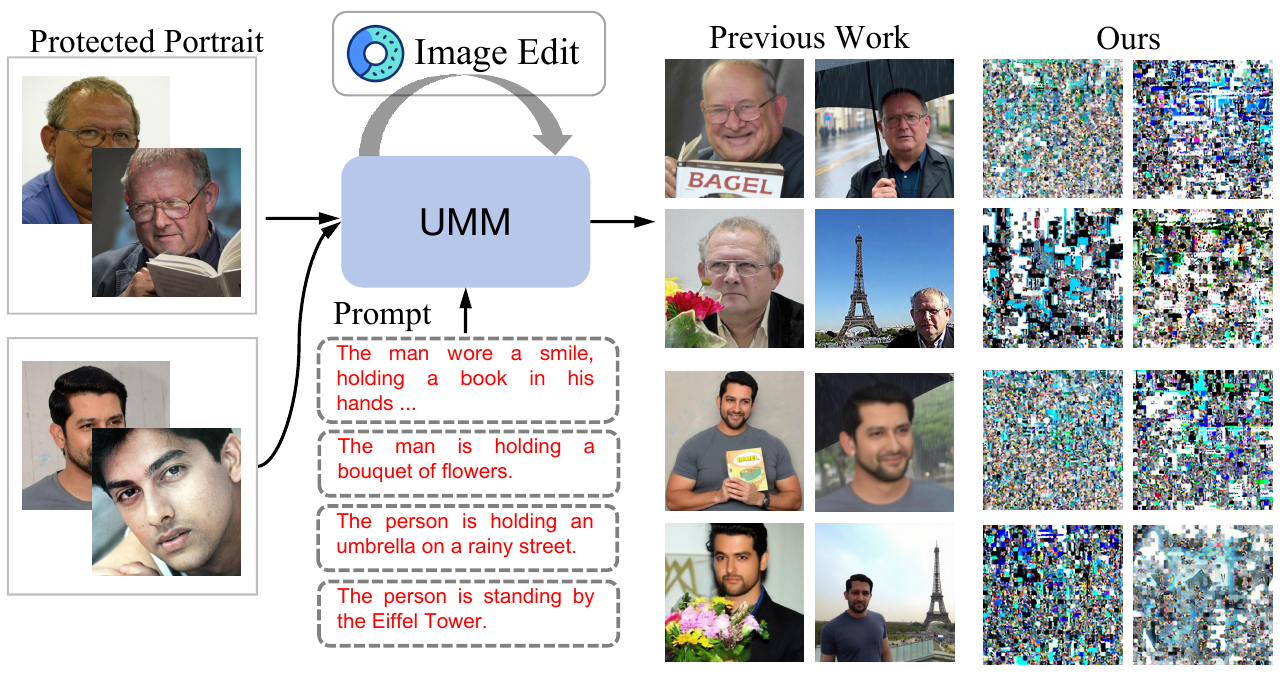}
    \caption{Overview of our proposed protection compared with previous methods~\cite{liang2023mist,mei2025veattack}.}
    \label{fig:framework}
\end{figure}

Currently, adversarial examples~\cite{Szegedy2013IntriguingPO, Goodfellow2014ExplainingAH} offer a promising method to protect copyright. Unlike watermarking methods~\cite{Li2024TowardsRV} that trace misuse after infringement has occurred, adversarial protection adds very small perturbations to an image before its release. They interfere directly with the model and prevents it from learning or reproducing protected content. Existing methods mainly target diffusion models or VLMs. Diffusion-based methods often perturb the VAE latent space or the predictions of denoising network~\cite{shan2023glaze, liang2023mist, liang2023adversarial, li2024pid, yang2025variance}. VLM-based methods mainly change the semantic features produced by a visual encoder~\cite{mei2025veattack}. These methods have shown promising results separately. However, the editing process of modern UMMs often depends on several visual branches at the same time. It remains unclear whether protection methods designed for a single branch can work in this setting. 

We study this problem in UMMs with separate visual encoders: a ViT for understanding and a VAE for generation. As shown in Figure~\ref{fig:framework}, we adapt diffusion-based protection methods and VLM-based adversarial methods to this UMM setting. However, neither group can reliably prevent facial editing. In fact, diffusion-based methods mainly perturb the VAE branch, while the ViT branch still provides a stable semantic representation of the face. In contrast, VLM-based ones mainly perturb the ViT branch, while the VAE branch retains useful appearance and identity information. Therefore, the two branches can compensate for each other. As long as either branch preserves a reliable facial representation, the UMM may still complete the requested edit.

To further understand this failure, we trace how ViT-only and VAE-only protection methods affect the two visual encoders, the shared Mixture-of-Transformers (MoT) tokens, and the final generation head. Although existing methods cause clear feature changes within their target branches, these local changes do not always propagate to the final generation process. This suggests that perturbing an individual branch is insufficient. We then argue that effective protection should satisfy two conditions. First, both branches should deviate from their clean representations, preventing either branch from serving as a stable fallback. Second, the two branches should provide conflicting representations of the same facial identity, disrupting their cooperation during multimodal fusion. Existing methods partially satisfy the first condition but retain residual cross-branch consistency. This observation motivates our method design, which jointly perturbs both branches and \textbf{directly} breaks their representation agreement.

Based on this, we propose \textbf{C}ross-Branch \textbf{C}onflict as a \textbf{S}hield (\textbf{CCS}), a proactive protection method against facial editing of UMMs. CCS applies untargeted divergence objectives to the ViT and VAE tokens in the shared language-model input space, preventing either branch from retaining a stable representation of the original face. It further introduces a consistency disruption objective based on linear Centered Kernel Alignment (CKA) to reduce the structural agreement between the protected ViT and VAE tokens, producing conflicting representations of the same input. The branch-wise objectives disrupt useful information in each path, while the cross-branch objective prevents the model from combining the remaining information into a consistent facial representation. During optimization, all model parameters remain frozen, and CCS only updates a bounded input perturbation. 

To evaluate CCS, we adopt a comprehensive protocol covering identity preservation, instruction completion, and image quality. Extensive experiments show that CCS preserves input fidelity while substantially reducing output identity similarity and editing completion.

Our main contributions are summarized as follows:
\begin{itemize}
\item We first systematically study facial identity protection for unified multimodal editing. We identify a shared limitation of existing  methods: they perturb single visual branch and leave the cross-branch cooperation largely intact.
\item We propose CCS, which jointly optimizes ViT-branch divergence, VAE-branch divergence, and CKA-based cross-branch consistency disruption. This design creates both representation shifts and branch conflicts in the shared token space, thereby hindering reliable feature extraction.
\item Extensive experiments show that CCS can reliably prevent effective facial editing while preserving the visual quality of the protected input images.
\end{itemize}
\section{Related Works}
\label{sec:related}

\textbf{Protective Perturbations against Image Misuse. } Recent works have explored adversarial examples as a proactive defense against unauthorized use of copyrighted or private images in diffusion models. Early methods such as AdvDM and Mist~\cite{liang2023adversarial,liang2023mist} perturb artworks before release, preventing diffusion models from faithfully imitating protected content or styles. Glaze~\cite{shan2023glaze} further protects artists by cloaking style representations learned by text-to-image models, while Nightshade~\cite{shan2024nightshade} extends this idea to data poisoning, corrupting model behavior when protected images are scraped for training. Recently, PID~\cite{li2024pid} relaxes the strong assumption that defenders know the attacker’s prompts by attacking prompt-independent visual representations, and VAC~\cite{yang2025variance} improves semantic erasure by optimizing variance-driven objectives in the latent space of customized diffusion models. Overall, these studies show that designed adversarial examples can alter how generative models encode protected data, thereby providing an effective technical route for copyright and privacy protection. 

\noindent \textbf{Unified multimodal models. } Unified multimodal models (UMMs) integrate the understanding and generation capabilities in one model. A variety of UMM architectures~\cite{chen2025janus,deng2025emerging,xie2026show,tian2026internvl,diao2026sensenova} have been developed. Among them, separate-encoder UMMs like BAGEL~\cite{deng2025emerging} and InternVL-U~\cite{tian2026internvl}, decouple the processing of two distinct types of representations (usually ViT and VAE), and have become a prominent architectural family. This is the main architectural family studied here.

Recent work~\cite{su2026unigame} demonstrates the jailbreak vulnerabilities of UMMs. However, few studies have explored safeguarding identities in the widely used aspect of UMMs: multimodal image editing. 

\begin{figure*}
    \centering
    \includegraphics[width=\linewidth]{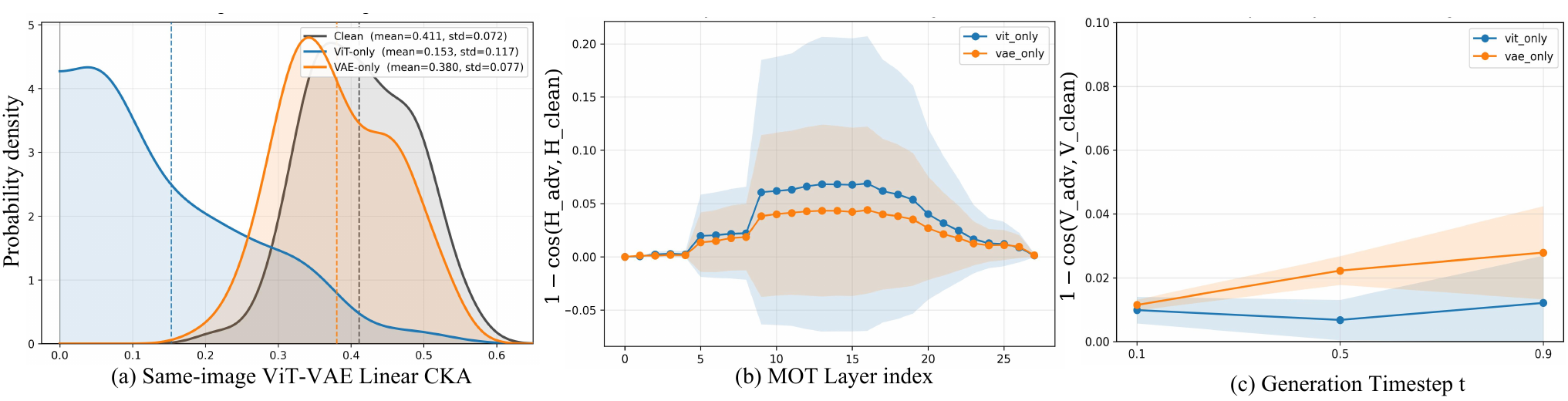}
    \caption{The three subfigures showing results of exploratory experiments. (a) The probability density on CKA of the representations extracted from the same image by the ViT and VAE encoders. (b) The similarity of features from two branches on different MoT layers. (c) Effects of protection methods on generation velocity across timesteps.}
    \label{fig:observation}
\end{figure*}


\section{Defensive Assumptions}
\label{sec:preliminary}
\textbf{Defense goal.} As mentioned in the Introduction, an unauthorized user can take a publicly available face image and use a UMM to generate edited images with the identity of the person. Such unauthorized facial editing can lead to portrait misuse, or the creation of unwanted facial derivatives. In this work, we focus on proactive protection against such misuse. Specifically, the defender aims to add a small perturbation to the facial image before releasing it. When an unauthorized user later applies a UMM to edit this protected image, the hidden perturbation disrupts the facial representations used by the model, making it difficult to preserve the original identity in the generated result. There exist some constraints: (1) the perturbation should be \textbf{unnoticeable} to humans, (2) it should disturb the UMM editing process and \textbf{prevent} the model from producing identity-preserving edited results. 

\noindent \textbf{Defender capabilities.} We assume a grey-box setting, that the defender can only access the visual encoding modules of the target UMM, or a surrogate UMM with a similar visual encoding design. This assumption is similar to previous settings in generating adversarial examples for VLMs~\cite{wang2024break, wang2024transferable, yin2023vlattack}. During optimization, all model parameters are non-editable to the defender. Only the input perturbation is updated under a bounded budget. The protected image can then be released publicly. 
\section{Methodology}
\label{sec:method}

\begin{figure*}
    \centering
    \includegraphics[width=\linewidth]{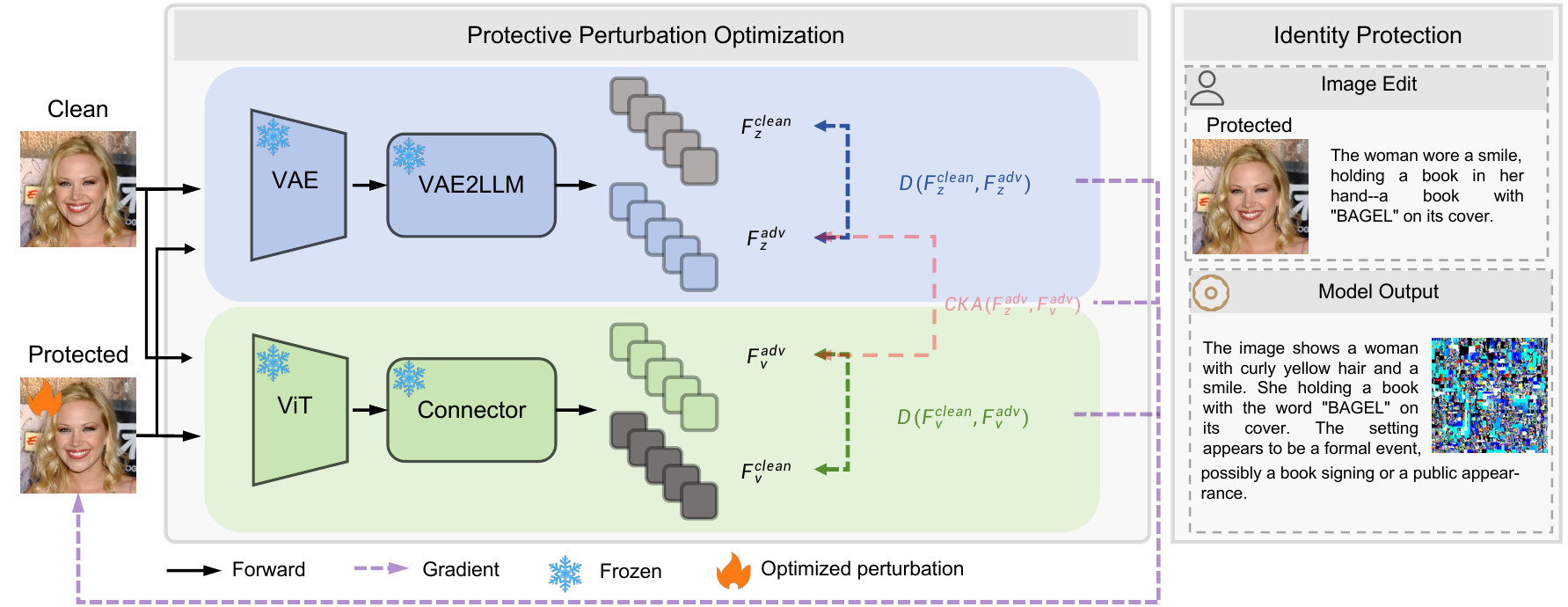}
    \caption{Overview of our method CCS. The `protected' image is with adversarial perturbation. Our method mainly consists of one cross-branch conflict part, and two branch-wise representation divergences. During optimization, these loss functions are combined together. The protected image can effectively disrupt UMM, leading to image-editing failures.}
    \label{fig:method}
\end{figure*}

\subsection{Problem Formulation}
\label{sec:formulation}

On having the defensive assumptions described above, we make the following problem formulation:

Given a clean image
$\mathbf{x}\in[0,1]^{H\times W\times 3}$,
we construct a protected image $\mathbf{x}^{p}$ as
\begin{equation}
    \mathbf{x}^{p}
    =
    \Pi_{[0,1]}
    \left(
        \mathbf{x}+\boldsymbol{\delta}
    \right),
    \qquad
    \|\boldsymbol{\delta}\|_{\infty}\leq\epsilon,
    \label{eq:protected_image}
\end{equation}
where $\boldsymbol{\delta}$ is the protective perturbation,
$\epsilon$ is the perturbation budget,
and $\Pi_{[0,1]}$ clips the image into the valid pixel range.

A UMM contains a generation branch and an understanding branch.
We denote their output features as
\begin{equation}
    \mathbf{F}_{z}(\mathbf{x})
    =
    \mathcal{G}_{z}(\mathbf{x}),
    \qquad
    \mathbf{F}_{v}(\mathbf{x})
    =
    \mathcal{G}_{v}(\mathbf{x}),
    \label{eq:branch_features}
\end{equation}
where $\mathcal{G}_{z}$ and $\mathcal{G}_{v}$ are the two frozen visual branches.
Their features are mapped into the shared LLM-input token space.
Thus, the two branches need to provide compatible representations of the
same image. Based on this property, we design the protection objective from two perspectives.
First, we reduce the agreement between the generation and understanding
branches.
Second, we enlarge the feature difference between the clean and protected
images within each branch.
The protective perturbation is obtained by
\begin{equation}
    \boldsymbol{\delta}^{*}
    =
    \arg\max_{\|\boldsymbol{\delta}\|_{\infty}\leq\epsilon}
    \mathcal{L}_{\mathrm{total}}
    \left(
        \mathbf{x},
        \mathbf{x}^{p}
    \right).
    \label{eq:general_objective}
\end{equation}
During optimization, only the input perturbation is updated. The whole process does not use any editing prompt.


\subsection{Observation: Cross-Branch Compatibility and Perturbation Attenuation}
We start with BAGEL-7B-MoT~\cite{deng2025emerging} to study the problem. BAGEL uses a ViT branch~\cite{Dosovitskiy2020AnII} for semantic understanding and a VAE branch~\cite{Kingma2013AutoEncodingVB} for texture-aware image generation. The outputs of both branches are mapped into a shared language-model input space and jointly processed by the following MoT layers~\cite{deng2025emerging}. We conduct exploratory experiments to investigate whether this property can be exploited by our protection method. Clean, ViT-level protection~\cite{mei2025veattack} and VAE-level protection~\cite{yang2025variance} are utilized. Firstly we apply centered kernel alignment~\cite{cortes2012algorithms} (CKA) to check the representations $\mathbf{F}_{z}(\mathbf{x})$ and $\mathbf{F}_{v}(\mathbf{x})$ from the VAE and ViT encoders. For two centered feature matrices $\mathbf{X}$ and $\mathbf{Y}$, linear CKA is defined as
\begin{equation}
    \operatorname{CKA}(\mathbf{X},\mathbf{Y})
    =
    \frac{
        \left\|
            \mathbf{Y}^{\top}\mathbf{X}
        \right\|_{F}^{2}
    }{
        \left\|
            \mathbf{X}^{\top}\mathbf{X}
        \right\|_{F}
        \left\|
            \mathbf{Y}^{\top}\mathbf{Y}
        \right\|_{F}
    },
    \label{eq:cka}
\end{equation}
where $\|\cdot\|_{F}$ denotes the Frobenius norm.
A larger CKA value means stronger agreement between the two feature spaces. Results are in Figure~\ref{fig:observation}(a). We discover that features from two branches exhibit structural agreement~\cite{li2026g,liu2026tuna}, while single-branch adversarial \textbf{perturbations reduce this cross-branch similarity} to different degrees. However, Figure~\ref{fig:observation}(b) suggests that these deviations are obvious in intermediate MoT hidden states, but gradually \textbf{diminish in later layers}. Figure~\ref{fig:observation}(c) further shows that the \textbf{protection effects are asymmetric} across timesteps, indicating that perturbing a single visual branch cannot consistently disrupt the entire unified editing process. These findings may explain why perturbation only on one branch fails to protect facial identities. This motivates us to consider a method that breaks the agreement above directly.

\subsection{Cross-Branch Conflict}
\label{sec:cross_branch}

Figure~\ref{fig:method} shows the workflow of CCS. Regarding the observations above, our core idea is to directly disrupt the agreement to disturb the editing process. The two branches may produce different numbers of tokens.
We use a token mapping function $\Phi(\cdot)$ to convert them into the same token length:
\begin{equation}
    \widehat{\mathbf{F}}_{b}(\mathbf{x})
    =
    \Phi
    \left(
        \mathbf{F}_{b}(\mathbf{x})
    \right),
    \qquad
    b\in\{z,v\}.
    \label{eq:token_mapping}
\end{equation}
When the two branches already produce the same number of tokens,
$\Phi(\cdot)$ is an identity mapping. Details about the mapping function and linear CKA are shown in Appendix C.

The two encoders have different structures and feature distributions.
Thus, direct element-wise comparison is not suitable.
Inspired from the study on feature changes, we also use 
CKA to measure the relation between their
features.

We define the cross-branch agreement of an image as
\begin{equation}
    \mathcal{A}(\mathbf{x})
    =
    \operatorname{CKA}
    \left(
        \widehat{\mathbf{F}}_{z}(\mathbf{x}),
        \widehat{\mathbf{F}}_{v}(\mathbf{x})
    \right).
    \label{eq:branch_alignment}
\end{equation}

Having this, we consider a new cross-branch conflict loss: 
\begin{equation}
    \mathcal{L}_{\mathrm{cbc}}
    =
    \mathcal{A}(\mathbf{x})
    -
    \mathcal{A}(\mathbf{x}^{p}).
    \label{eq:cbc_loss}
\end{equation}

The first term measures the original agreement between the two branches.
It remains fixed during optimization.
The second term measures their agreement after protection.
Maximizing $\mathcal{L}_{\mathrm{cbc}}$ reduces the CKA value of the
protected image.

As a result, the understanding branch and the generation branch form
conflicting representations of the same input.
The shared language model then receives an unstable visual context.
This makes it harder for the UMM to connect the editing instruction with the
correct visual content.

\subsection{Branch-wise Representation Divergence}
\label{sec:branch_divergence}

The cross-branch objective reduces the agreement between the two branches.
However, a lower agreement does not ensure that both branches are changed.
For example, one branch may shift greatly, while the other branch still
keeps useful information about the clean image.
We therefore add two branch-wise objectives.

We use a normalized feature distance
$\mathcal{D}(\cdot,\cdot)$ to measure the difference between clean and
protected features.
For two token sequences
$\mathbf{X},\mathbf{Y}\in\mathbb{R}^{N\times d}$,
we define
\begin{equation}
    \mathcal{D}(\mathbf{X},\mathbf{Y})
    =
    1-
    \frac{1}{N}
    \sum_{i=1}^{N}
    \frac{
        \mathbf{x}_{i}^{\top}\mathbf{y}_{i}
    }{
        \|\mathbf{x}_{i}\|_{2}
        \|\mathbf{y}_{i}\|_{2}
    }.
    \label{eq:feature_distance}
\end{equation}

\paragraph{VAE Branch Divergence.}
The generation branch uses a VAE encoder to keep the visual information
needed for reconstruction and synthesis.
We define its divergence loss as
\begin{equation}
    \mathcal{L}_{\mathrm{vae}}
    =
    \mathcal{D}
    \left(
        \mathbf{F}_{z}(\mathbf{x}),
        \mathbf{F}_{z}(\mathbf{x}^{p})
    \right).
    \label{eq:vae_loss}
\end{equation}

Maximizing this loss moves the protected VAE features away from the clean features. This weakens the representation of appearance and structure used in later image generation.

\paragraph{ViT Branch Divergence.}
The understanding branch uses a ViT encoder to extract high-level visual semantics.
These features support object recognition, attribute understanding, and
instruction grounding.
We define its divergence loss as
\begin{equation}
    \mathcal{L}_{\mathrm{vit}}
    =
    \mathcal{D}
    \left(
        \mathbf{F}_{v}(\mathbf{x}),
        \mathbf{F}_{v}(\mathbf{x}^{p})
    \right).
    \label{eq:vit_loss}
\end{equation}

Maximizing this loss changes the semantic representation of the protected
image.
The model may then fail to identify the correct object mentioned in the editing prompt.

\begin{table*}[t]
    \centering

    \begin{minipage}[t]{0.49\textwidth}
        \centering
        \setlength{\tabcolsep}{3pt}
        \resizebox{\linewidth}{!}{
        \begin{tabular}{lcccc}
            \toprule
            \multirow{2}{*}{\textbf{Methods}} & \multicolumn{4}{c}{\textbf{Prompt~1}} \\ 
            \cmidrule(lr){2-5}
            & \textbf{ISM($\downarrow$)} & \textbf{FID($\uparrow$)} & \textbf{BRISQUE($\uparrow$)} & \textbf{Und-Score($\downarrow$)} \\ 
            \midrule
            Clean       & 0.48 & 100.46 & 26.45 & 100.00 \\ 
            \midrule 
            AdvDM       & 0.36 & 169.38 & 29.53 & 87.75 \\ 
            PhotoGuard  & 0.20 & 196.25 & 30.53 & 79.75 \\ 
            Mist        & 0.19 & 245.05 & 36.13 & 64.25 \\ 
            PID         & 0.17 & 291.69 & 44.35 & 70.00 \\ 
            VAC         & 0.32 & 119.79 & 23.48 & 96.50 \\ 
            VeAttack    & 0.29 & 122.15 & 22.34 & 98.50 \\ 
            \textbf{CCS (ours)} & \textbf{0.00} & \textbf{532.38} & \textbf{46.49} & \textbf{0.00} \\ 
            \bottomrule
        \end{tabular}
        }
        \captionof{table}{Performance comparison of different methods on prompt 1: ``The person wore a smile, holding a book in hands-with `BAGEL' on its cover''.}
        \label{tab:method_comparison}
    \end{minipage}
    \hfill
    \begin{minipage}[t]{0.49\textwidth}
        \centering
        \setlength{\tabcolsep}{3pt}
        \resizebox{\linewidth}{!}{
        \begin{tabular}{lcccc}
            \toprule
            \multirow{2}{*}{\textbf{Methods}} & \multicolumn{4}{c}{\textbf{Prompt 2}} \\ 
            \cmidrule(lr){2-5}
            & \textbf{ISM($\downarrow$)} & \textbf{FID($\uparrow$)} & \textbf{BRISQUE($\uparrow$)} & \textbf{Und-Score($\downarrow$)} \\ 
            \midrule
            Clean       & 0.15 & 58.84 & 20.13 & 96.88 \\ 
            \midrule 
            AdvDM       & 0.11 & 96.17 & 25.05 & 87.50 \\ 
            PhotoGuard  & 0.07 & 133.86 & 32.92 & 78.12 \\ 
            Mist        & 0.05 & 179.06 & 38.50 & 65.31 \\ 
            PID         & 0.08 & 171.90 & 42.00 & 72.19 \\ 
            VAC         & 0.07 & 108.21 & 29.94 & 92.50 \\ 
            VeAttack    & 0.05 & 68.33 & 19.73 & 95.67 \\ 
            \textbf{CCS (ours)} & \textbf{0.00} & \textbf{437.74} & \textbf{44.34} & \textbf{0.04} \\ 
            \bottomrule
        \end{tabular}
        }
        \captionof{table}{Performance comparison of different methods on prompt 2: ``The person is holding an umbrella on a rainy street''.}
        \label{tab:method_comparison_p2}
    \end{minipage}

    \vspace{0.8em}

    \begin{minipage}[t]{0.49\textwidth}
        \centering
        \setlength{\tabcolsep}{3pt}
        \resizebox{\linewidth}{!}{
        \begin{tabular}{lcccc}
            \toprule
            \multirow{2}{*}{\textbf{Methods}} & \multicolumn{4}{c}{\textbf{Prompt~3}} \\ 
            \cmidrule(lr){2-5}
            & \textbf{ISM($\downarrow$)} & \textbf{FID($\uparrow$)} & \textbf{BRISQUE($\uparrow$)} & \textbf{Und-Score($\downarrow$)} \\ 
            \midrule
            Clean       & 0.68 & 87.36 & 11.01 & 97.92 \\ 
            \midrule 
            AdvDM       & 0.49 & 125.70 & 14.46 & 90.00 \\ 
            PhotoGuard  & 0.30 & 153.61 & 19.16 & 81.67 \\ 
            Mist        & 0.29 & 192.27 & 31.42 & 64.58 \\ 
            PID         & 0.23 & 195.13 & 39.03 & 70.83 \\ 
            VAC         & 0.52 & 101.27 & 24.64 & 90.30 \\ 
            VeAttack    & 0.58 & 90.54 & 21.52 & 96.67 \\ 
            \textbf{CCS (ours)} & \textbf{0.00} & \textbf{471.32} & \textbf{45.32} & \textbf{6.67} \\ 
            \bottomrule
        \end{tabular}
        }
        \captionof{table}{Performance comparison of different methods on Prompt~3: ``The person is holding a bouquet of flowers''.}
        \label{tab:method_comparison_p3}
    \end{minipage}
    \hfill
    \begin{minipage}[t]{0.49\textwidth}
        \centering
        \setlength{\tabcolsep}{3pt}
        \resizebox{\linewidth}{!}{
        \begin{tabular}{lcccc}
            \toprule
            \multirow{2}{*}{\textbf{Methods}} & \multicolumn{4}{c}{\textbf{Prompt~4}} \\ 
            \cmidrule(lr){2-5}
            & \textbf{ISM($\downarrow$)} & \textbf{FID($\uparrow$)} & \textbf{BRISQUE($\uparrow$)} & \textbf{Und-Score($\downarrow$)} \\ 
            \midrule
            Clean       & 0.15 & 80.35 & 24.59 & 85.83 \\ 
            \midrule 
            AdvDM       & 0.09 & 108.06 & 27.80 & 73.75 \\ 
            PhotoGuard  & 0.03 & 163.78 & \textbf{43.42} & 42.50 \\ 
            Mist        & 0.05 & 199.63 & 43.14 & 41.25 \\ 
            PID         & 0.02 & 142.11 & 34.16 & 64.58 \\ 
            VAC         & 0.14 & 87.57 & 29.08 & 79.58 \\ 
            VeAttack    & 0.10 & 97.46 & 28.89 & 76.67 \\ 
            \textbf{CCS (ours)} & \textbf{0.00} & \textbf{423.83} & 37.30 & \textbf{2.50} \\ 
            \bottomrule
        \end{tabular}
        }
        \captionof{table}{Performance comparison of different methods on Prompt~4: ``The person is standing by the Eiffel Tower''.}
    \label{tab:method_comparison_p4}
    \end{minipage}

\end{table*}

\subsection{Overall Objective and Optimization}
\label{sec:optimization}

We combine the cross-branch conflict loss and the two branch-wise losses:
\begin{equation}
    \mathcal{L}_{\mathrm{total}}
    =
    \alpha\mathcal{L}_{\mathrm{vae}}
    +
    \beta\mathcal{L}_{\mathrm{vit}}
    +
    \gamma\mathcal{L}_{\mathrm{cbc}},
    \label{eq:total_loss}
\end{equation}
where $\alpha$, $\beta$, and $\gamma$ control the weights of the three
objectives.

The protected image is obtained by solving
\begin{equation}
    \mathbf{x}^{p*}
    =
    \arg\max_{
        \|\mathbf{x}^{p}-\mathbf{x}\|_{\infty}\leq\epsilon
    }
    \mathcal{L}_{\mathrm{total}}.
    \label{eq:final_objective}
\end{equation}

We use projected gradient ascent to update the protected image ($t$ means a certain step). where $\eta$ is the step size and
$\mathcal{B}_{\infty}(\mathbf{x},\epsilon)$ is the
$\ell_{\infty}$ neighborhood of the clean image.
\begin{equation}
    \mathbf{x}^{p}_{t+1}
    =
    \Pi_{
        \mathcal{B}_{\infty}(\mathbf{x},\epsilon)
        \cap [0,1]
    }
    \left[
        \mathbf{x}^{p}_{t}
        +
        \eta\,
        \operatorname{sign}
        \left(
            \nabla_{\mathbf{x}^{p}_{t}}
            \mathcal{L}_{\mathrm{total}}
        \right)
    \right],
    \label{eq:pgd_update}
\end{equation}

\begin{figure*}[t]
    \centering
    \includegraphics[width=\linewidth]{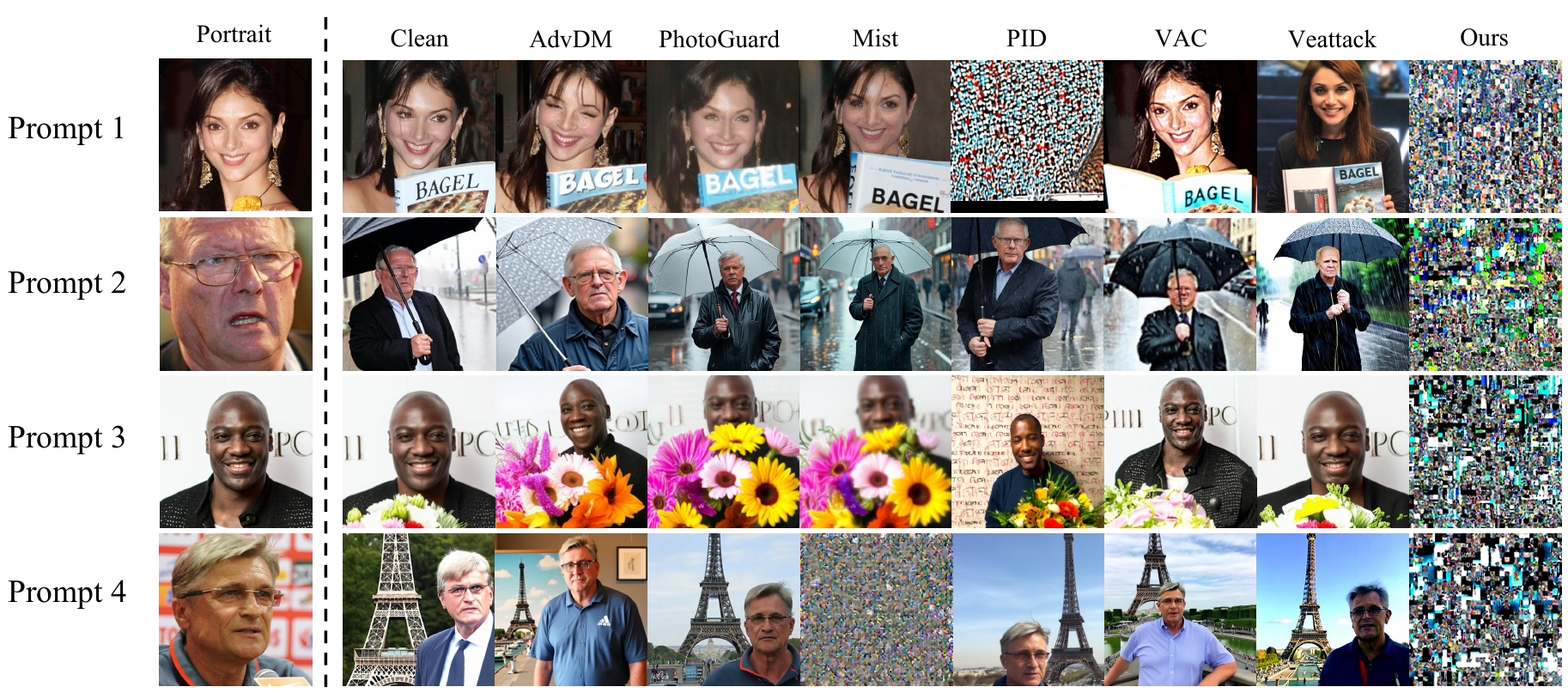}
    \caption{Qualitative results of our method v.s. baselines. CCS demonstrates substantially stronger protection. }
    \label{fig:main_results}
\end{figure*}

\section{Experiments}
\label{sec:experiments}

\subsection{Experimental Settings}
\textbf{Datasets and Models.}
Our experiments are mainly conducted on VGGFace2~\cite{Cao2017VGGFace2AD}, where we select 20 different facial identities (four images per identity). We use BAGEL-7B as the surrogate model. Additional experiments on face datasets, such as CelebA-HQ~\cite{Karras2017ProgressiveGO} are detailed in Appendix F.

\noindent \textbf{Baselines.} We choose diffusion-based and VLM-based baselines like AdvDM~\cite{liang2023adversarial}, PhotoGuard~\cite{salman2023raising}, Mist~\cite{liang2023mist}, PID~\cite{li2024pid}, VAC~\cite{yang2025variance}, and VeAttack~\cite{mei2025veattack}. Details of baselines can also be found in Appendix D.

\noindent \textbf{Settings.} We conduct our experiments on NVIDIA A100 80G GPUs. All adversarial examples are optimized for 100 iterations, with the perturbation budget set to 8/255. Unless otherwise specified, all baselines use their default hyperparameters. For our method, the loss weights $\alpha$, $\beta$, and $\gamma$ are set to 10, 0.2, and 1.0. The interpolated spatial resolution in the mapping function is set to $16 \times 16$. Further analyses of the hyper-parameters are provided in the Appendix G.

\noindent \textbf{Evaluation Metric.} Note that all the settings here are based on a defensive perspective. For evaluation, we assess our method from some perspectives. For facial identity protection, we use the identity score matching (ISM)~\cite{Le2023AntiDreamBoothPU} to measure the facial similarity between the generated image and the original image (lower is better). For output distribution disruption, we use FID~\cite{Heusel2017GANsTB} to measure the distributional discrepancy between images edited from adversarial examples and those edited from the corresponding clean images (higher is better). To quantify output-quality degradation, we report BRISQUE~\cite{6272356}, where a higher score indicates stronger distortion and hence stronger defensive disruption. Moreover, we introduce a new metric to evaluate the defensiveness against instruction following. CLIP-Scores~\cite{Hessel2021CLIPScoreAR} are too coarse-grained to capture fine-grained instruction consistency~\cite{Li2024ChemVLMET, Ma2024I2EBenchAC, Hui2024HQEditAH}. We introduce the Understanding Score (\textbf{Und-Score}), which decomposes each instruction into atomic requirements and uses the UMM’s understanding branch to verify whether each requirement is satisfied by the edited image. A lower Und-Score indicates weaker instruction adherence (lower is better). More details are provided in Appendix~B.

\subsection{Main Experiments}

We \textbf{bold} the best results for all the metrics. As shown in Tables~\ref{tab:method_comparison}-\ref{tab:method_comparison_p4}, CCS achieves the lowest ISM and Und-Score, as well as the highest FID, under all four editing prompts, indicating overall best performance. In particular, CCS reduces ISM to 0.00 in every setting, showing that the edited results no longer keep the facial identity of the source portrait. Averaged over the four prompts, CCS obtains an FID of 466.32 and an Und-Score of 2.30. In comparison, the strongest baseline results on these two metrics are only 204.00 and 58.85. CCS also achieves the highest average BRISQUE score of 43.36. Although PhotoGuard produces a higher BRISQUE score on Prompt~4, it retains more identity information and satisfies more instruction requirements. Overall, single-branch methods provide only partial protection, since the other visual branch can still offer useful semantic or appearance cues. By jointly shifting both branches and breaking their agreement, CCS provides stronger protection across different editing instructions. More experiments on non-facial images are provided in the Appendix I.

\subsection{Ablation Study}

We ablate the two main components of CCS on Prompt~1. As shown in Table~\ref{tab:ablation}, removing the branch-wise divergence losses causes a substantial performance drop. ISM increases from 0.00 to 0.37, FID decreases from 532.38 to 141.79, and the Und-Score rises to 96.50. In this setting, the CKA objective can reduce the structural agreement between the ViT and VAE branches, but it does not explicitly force either branch away from its clean representation. As a result, each branch may still preserve sufficient identity and visual information for the model to complete the edit. Removing the cross-branch conflict loss also weakens the protection, increasing ISM to 0.31 and Und-Score to 43.68. The branch-wise-only setting does not disrupt the agreement between the two perturbed branches. The two branches may therefore shift in a coordinated manner and remain mutually compatible in the shared token space, allowing the subsequent generation modules to partially recover and fuse the facial information. Combining both components produces the strongest protection. These results confirm that the two components address different failure modes and play complementary roles.

\begin{table}[t]
    \centering
    \setlength{\tabcolsep}{3pt}
    \resizebox{\linewidth}{!}{
    \begin{tabular}{lcccc}
        \toprule
        \textbf{Methods}
        & \textbf{ISM($\downarrow$)}
        & \textbf{FID($\uparrow$)}
        & \textbf{BRISQUE($\uparrow$)}
        & \textbf{Und-Score($\downarrow$)} \\
        \midrule
        Clean                & 0.48          & 100.46          & 26.45         & 100.00         \\
        \midrule
        \textbf{CCS (ours)}  & \textbf{0.00} & \textbf{532.38} & \textbf{46.49} & \textbf{0.00} \\
        w/o branch-wise      & 0.37          & 141.79          & 23.07          & 96.50          \\
        w/o cross-branch     & 0.31          & 281.29          & 42.70          & 43.68          \\
        \bottomrule
    \end{tabular}
    }
    \captionof{table}{Ablation study on branch-wise and cross-branch design. We choose prompt 1 as an example.}
    \label{tab:ablation}
\end{table}

\subsection{Qualitative Results}

\begin{figure}[t]
    \centering
    \begin{subfigure}[t]{0.49\columnwidth}
        \centering
        \includegraphics[
            width=\linewidth,
            trim=5 5 5 5,
            clip
        ]{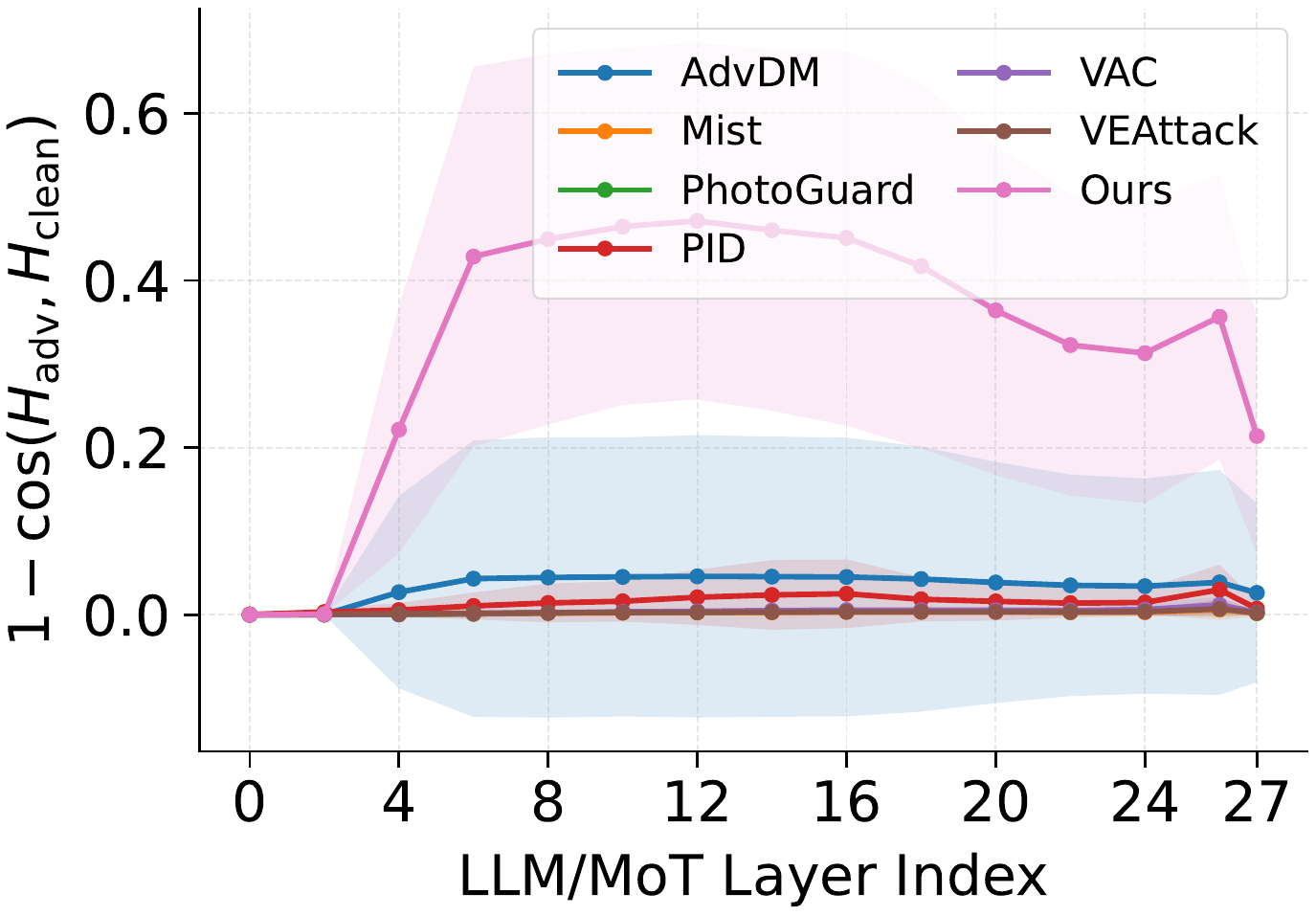}
        \caption{Layer hidden divergence.}
        \label{fig:hidden_divergence}
    \end{subfigure}
    \hfill
    \begin{subfigure}[t]{0.49\columnwidth}
        \centering
        \includegraphics[
            width=\linewidth,
            trim=5 5 5 5,
            clip
        ]{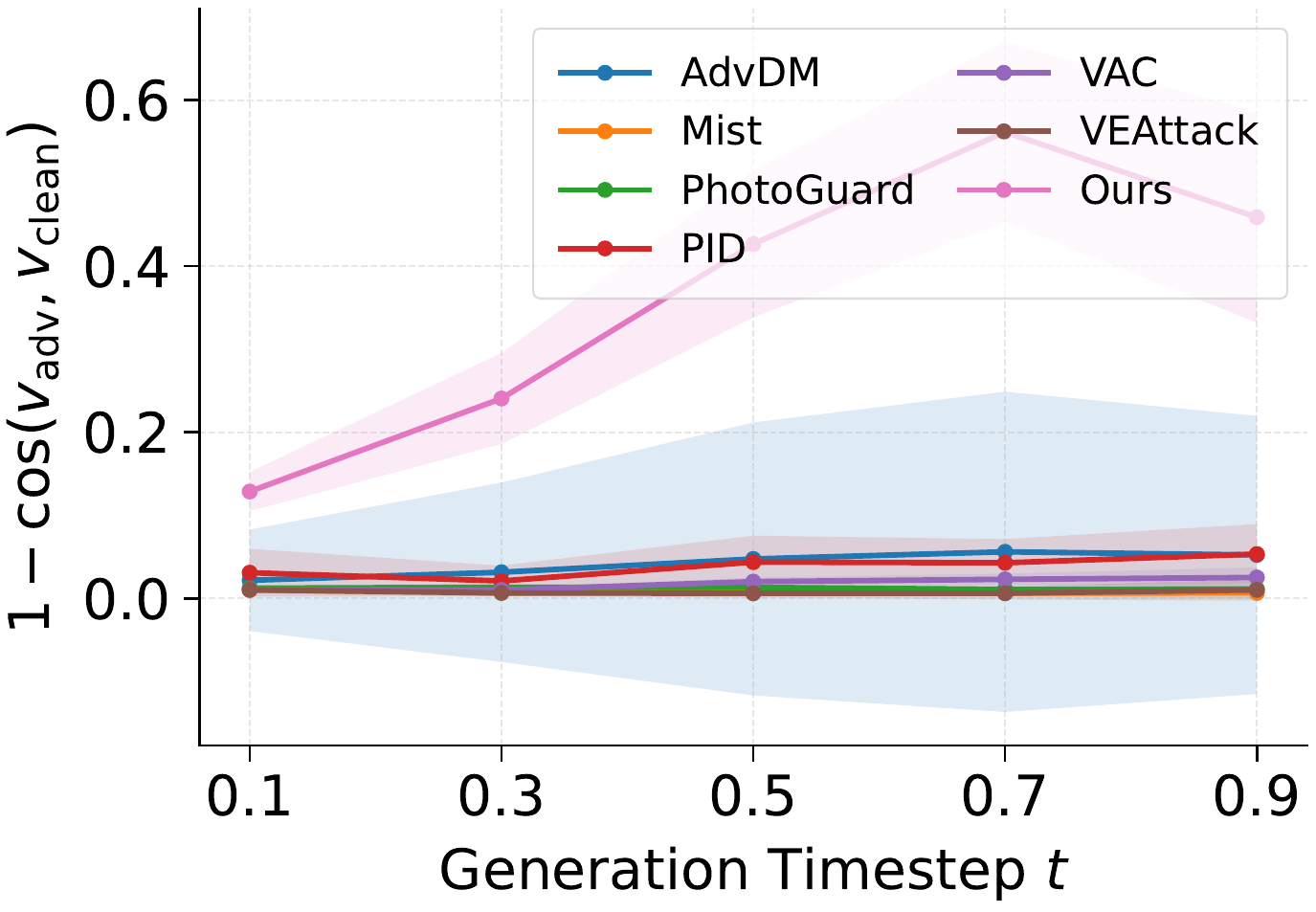}
        \caption{One-step velocity divergence.}
        \label{fig:velocity_divergence}
    \end{subfigure}

    \caption{
        Divergence induced by different adversarial protection methods.
        (a) Layer-wise divergence of generation hidden states.
        (b) One-step velocity prediction divergence across generation timesteps.
    }
    \label{fig:hidden_velocity_divergence}
\end{figure}

Figure~\ref{fig:main_results} presents qualitative comparisons under four editing instructions. Most baselines cause only limited or unstable disruption. In many cases, the edited images still preserve the source identity and contain the requested objects. Some methods cause generation failure in isolated cases, such as PID on Prompt~1 and Mist on Prompt~4. However, their protection is not consistent across different instructions. In contrast, CCS prevents identity-preserving editing. Its outputs contain severe visual artifacts or no longer resemble the source person. Most requested details are also removed or incorrectly generated. These results show that CCS provides more reliable visual protection than single-branch baselines. We further plot the divergence of hidden layers and the predicted flow velocity in~\Cref{fig:hidden_divergence,fig:velocity_divergence}. Compared with existing baselines, CCS induces substantially larger and more persistent shifts in both internal representations and velocity predictions. And this leads to stronger protection.

\subsection{Robustness Analysis}

We apply three common anti-protection methods, as also used in previous works, to study the robustness of CCS. These methods are image resizing, adding JPEG compression and Gaussian blur. They simulate the operations that unauthorized users may perform when aware of potential adversarial perturbations. CCS remains effective under all three transformations. It consistently reduces identity similarity, instruction-following performance, while substantially degrading editing quality compared with clean images.
\begin{table}[h]
    \centering
    \setlength{\tabcolsep}{3pt}
    \resizebox{\linewidth}{!}{
    \begin{tabular}{lcccc}
        \toprule
        \textbf{Methods}
        & \textbf{ISM($\downarrow$)}
        & \textbf{FID($\uparrow$)}
        & \textbf{BRISQUE($\uparrow$)}
        & \textbf{Und-Score($\downarrow$)} \\
        \midrule
        Clean         & 0.48 & 100.46 & 26.45 & 100.00 \\
        \midrule
        Resize        & 0.12 & 243.12 & 42.57 & 42.75 \\
        JPEG Comp.    & 0.30 & 179.07 & 34.57 & 70.75 \\
        Gaussian Blur & 0.00 & 351.99 & 39.33 & 11.75 \\
        \bottomrule
    \end{tabular}
    }
    \caption{Detailed results on robustness analysis of CCS. We choose prompt 1 as an example. Comp. means compression.}
    \label{tab:robustness}
\end{table}

\subsection{Additional Studies}
\begin{table}[t]
    \centering
    \setlength{\tabcolsep}{3pt}
    \resizebox{\linewidth}{!}{
    \begin{tabular}{lcccc}
        \toprule
        \textbf{Methods}
        & \textbf{ISM($\downarrow$)}
        & \textbf{FID($\uparrow$)}
        & \textbf{BRISQUE($\uparrow$)}
        & \textbf{Und-Score($\downarrow$)} \\
        \midrule
        Clean w/ think        & 0.23 & 80.41  & 23.19 & 98.69 \\
        \midrule
        \textbf{CCS} w/ think & 0.00 & 428.25 & 42.97 & 0.00  \\
        \bottomrule
    \end{tabular}
    }
    \captionof{table}{The results when opening thinking mode of BAGEL. We choose prompt 2 as an example.}
    \label{tab:thinking_mode}
\end{table}
\textbf{Thinking mode.} To study whether CCS still works when applying thinking-based editing, we open the thinking mode of BAGEL here. Results are in Table~\ref{tab:thinking_mode}. Although the clean images are more likely to be edited successfully under the thinking mode, as indicated by the improved editing metrics, our method still effectively protects facial identity information. More understanding-side prior (thinking mode) may not override the protection mechanism.


\begin{figure}[t]
    \centering
    \includegraphics[width=0.99\columnwidth]{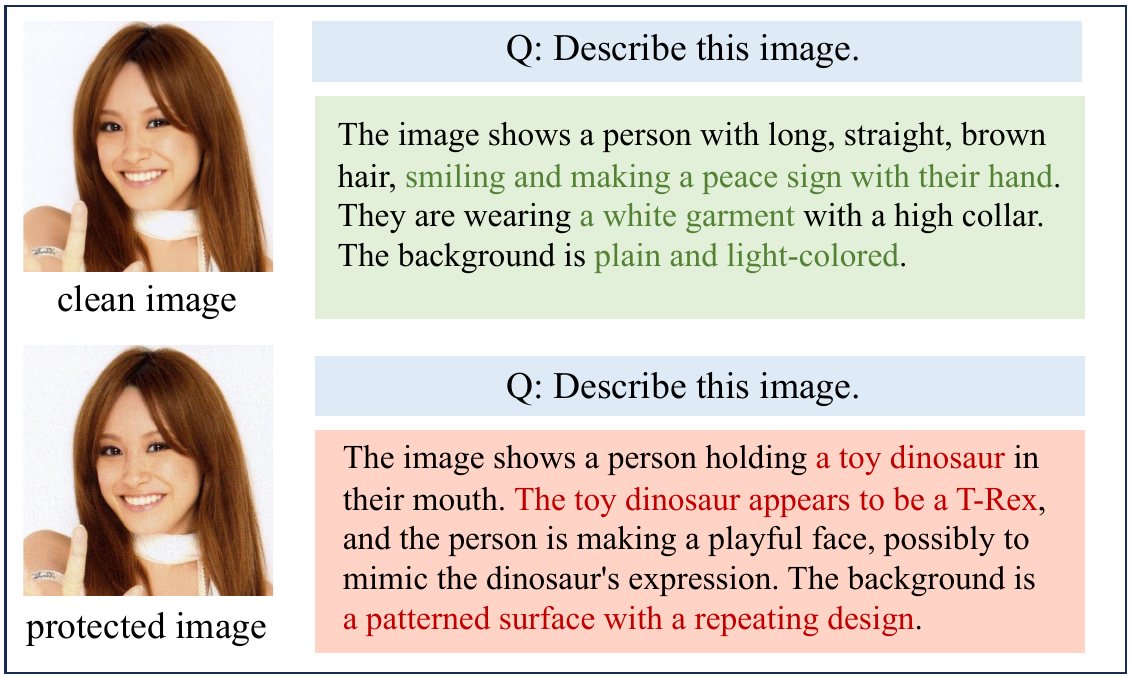}
    \caption{Understanding-side results. The adversarial perturbation is optimized using CCS. }
    \label{fig:und_adv}
\end{figure}
\begin{figure}[]
    \centering
    \includegraphics[width=0.85\columnwidth]{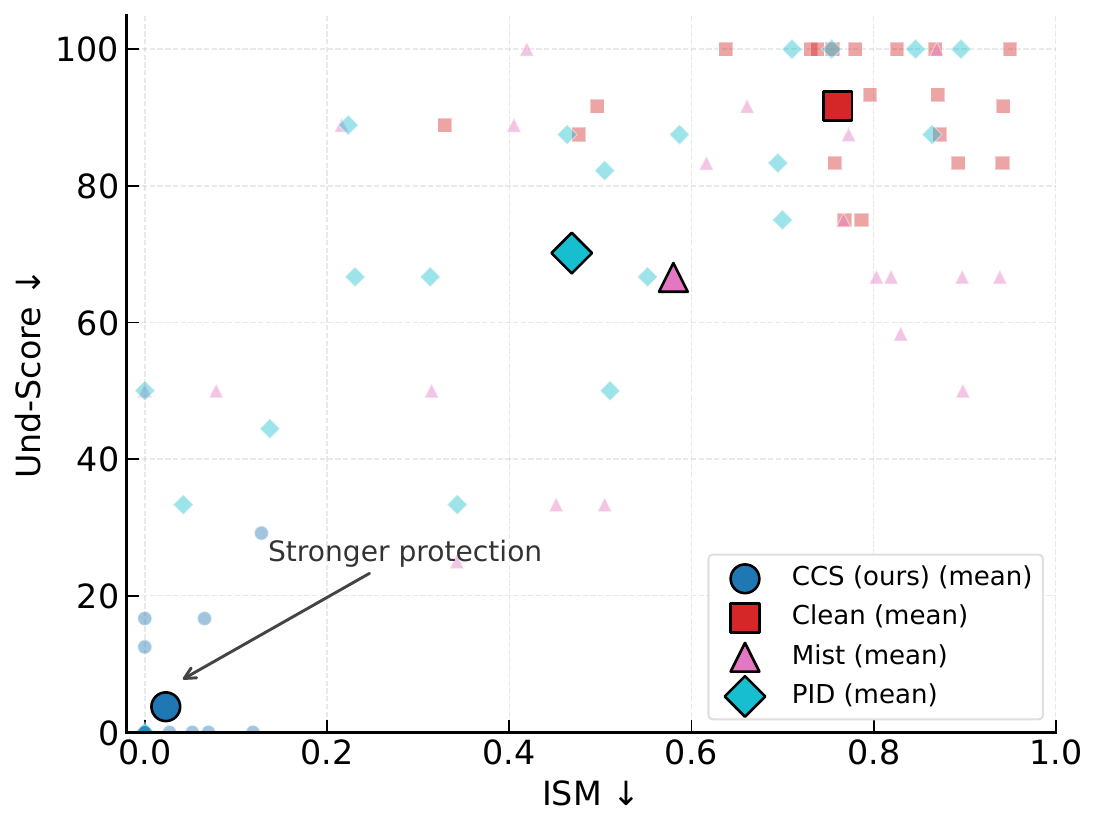}
    \caption{Protection performance across diverse editing prompts. CCS consistently provides strong protection.}
    \label{fig:prompt_agnostic}
\end{figure}
\noindent \textbf{Understanding-level performance.} Since UMMs also have capabilities on image understanding, we conduct an experiment to see if the understanding side can be perturbed, like Figure~\ref{fig:und_adv} depicts. The image protected by CCS leads the understanding side of BAGEL to generate wrong descriptions like ``toy dinosaur'' and ``patterned surface''. This suggests that disrupting feature agreement can also defend the improper understanding-level utilization of UMMs. 

\noindent \textbf{Prompt-Agnostic Protection.}
To evaluate sensitivity to editing instructions, we randomly generate 50 diverse prompts and use them to edit facial images, averaging the results across prompts for each identity. As shown in Figure~\ref{fig:prompt_agnostic}, CCS consistently places different identities near the bottom-left corner, achieving low ISM and Und-Score across diverse instructions. These results demonstrate that CCS provides stable protection without relying on specific editing prompts, supporting its prompt-agnostic nature.

\section{Conclusion}
\label{sec:conclusion}

In this work, we studied proactive facial identity protection for unified multimodal image editing. We found that single-branch protection is insufficient, since a UMM can still recover useful information from the other visual pathway. Based on this finding, we proposed CCS, which jointly shifts the ViT and VAE representations and disrupts their cross-branch agreement. Extensive experiments show that CCS consistently prevents identity-preserving editing and reduces instruction completion across different editing prompts. These results demonstrate the importance of cross-branch cooperation for UMMs and provide a promising direction for protecting facial images against unauthorized UMM editing.

\bibliography{aaai2027}

\clearpage

\urlstyle{rm}
\def\UrlFont{\rm}
\frenchspacing

\pdfinfo{
/TemplateVersion (2027.1)
}

\setcounter{secnumdepth}{2}

\title{Cross-Branch Conflict as a Shield: Safeguarding Facial Identities \\ in Unified Multimodal Image Editing \\ Technical Appendix}
\author{Anonymous Submission}
\affiliations{}


\maketitle

\appendix

\newpage 
\section{All Notations Used}

Here we list all the notations used in the main text in Table~\ref{tab:notations}.

\begin{table}[h]
        \centering
        \setlength{\tabcolsep}{3pt}
        \begin{tabular}{lc}
            \toprule
            Notations & Meanings \\ 
            \midrule
            $\mathbf{x}\in[0,1]^{H\times W\times 3}$      &  the clean image \\  
            $\mathbf{x}^{p}$ & the protected image \\ 
            $\boldsymbol{\delta}$  & protective perturbation \\ 
            $\epsilon$  & perturbation budget \\ 
            $\Pi_{[0,1]}$ & a clipping function \\
            $\mathcal{G}_{z}, \mathcal{G}_{v}$ & the ViT/VAE encoder \\
            $\mathbf{F}_{z}, \mathbf{F}_{v}$ & features from encoders \\
            $\mathcal{L}_{\mathrm{total}}$ & total optimization loss \\
            $\|\cdot\|_{F}$ & the Frobenius norm \\
            $\Phi(\cdot)$   & token mapping function \\
            $\widehat{\mathbf{F}}_{b}(\mathbf{x}), b\in\{z,v\} $ & feature after mapping \\
            $\mathcal{A}(\mathbf{\cdot})$ & cross-branch agreement \\
            $\mathcal{D}(\cdot,\cdot)$ & normalized feature distance \\
            $\mathcal{L}_{\mathrm{vit}}, \mathcal{L}_{\mathrm{vae}}, \mathcal{L}_{\mathrm{cbc}}$ & losses for optimization \\
            $\alpha, \beta, \gamma$ & hyper-parameters \\
            \bottomrule
        \end{tabular}
        
        \captionof{table}{All the notations we used in the main text.}
        \label{tab:notations}
\end{table}

\section{Und-Score}
Apart from traditional metrics, there should be one core metric evaluating whether the edited image follows the requirements mentioned in the instructions. Existing CLIP-based scores measure the overall semantic consistency between an image and a text. However, CLIP-based scores are too coarse-grained to capture fine-grained instruction consistency. Note that in our scenario, a better defense means that the UMM follows fewer requirements, representing a failed editing process. An intuitive way is to split the instructions into sub points, and evaluate them one by one. Based on this idea, we introduce the Understanding Score (\textbf{Und-Score}). Its motivation is that the understanding branch of a UMM can also interpret the image produced by its generation branch. We therefore use the understanding branch of a clean UMM to measure how well the edited image follows the instruction. Specifically, we first use Qwen2.5-7B-Instruct to decompose the prompt into a set of atomic scoring points. A special template is used to reduce overlap and ensure that all key requirements are covered. Here is the template:

\begin{tcolorbox}[
    colback=gray!10,
    colframe=gray!10,
    boxrule=0pt,
    arc=0pt
]
\noindent \textbf{System Prompt:} 

You are an evaluation-question generator for image editing. Your job is to decompose an editing prompt into atomic, visually checkable yes/no questions. Each question should correspond to exactly one key visual requirement.

\noindent \textbf{User Prompt:} 

Original editing prompt:
\{prompt\}

Please decompose it into atomic yes/no questions. Do not include very similar questions.

Rules:

1. One key visual requirement -> one yes/no question.

2. The correct answer should be ``Yes'' if the edited image satisfies the prompt.

3. Ask questions that can be answered from the final edited image alone.

4. For removal or negative requirements, write the question so that ``Yes'' means the requirement is satisfied.

   Example: ``remove the dog'' -> ``Is there no dog in the image?''
   
5. Avoid subjective questions unless the visual attribute is clear.

6. Output JSON only, no markdown, no explanation.

7. The JSON format should be correct. There shouldn't be a``,'' at the end of the last sentence.

Output format:

[
  \{\{``id'': 1, ``question'': ``Is there a cat in the image?'', ``aspect'': ``cat exists''\}\},
  
  \{\{``id'': 2, ``question'': ``Is the cat wearing a red hat?'', ``aspect'': ``red hat on cat''\}\}
]

Example:

Prompt: ``Make the cat wear a red hat and put a blue cup on the table.''

Output:

[
  \{\{``id'': 1, ``question'': ``Is there a cat in the image?'', ``aspect'': ``cat exists''\}\},
  
  \{\{``id'': 2, ``question'': ``Is the cat wearing a red hat?'', ``aspect'': ``red hat on cat''\}\},
  
  \{\{``id'': 3, ``question'': ``Is there a blue cup on the table?'', ``aspect'': ``blue cup on table''\}\}
]
\end{tcolorbox}

The understanding branch of the UMM, BAGEL in our experiments, then checks each scoring point against the edited image and assigns a binary score of 0 or 1. Specifically, for each scoring point, we ask the understanding side of the UMM one yes/no question, like ``Is this xxx red?''. The given edited image is fed into the UMM for each yes/no question. Then, if one key point is satisfied, then assign 1. If not, assign 0. The final Und-Score is the average percentage score over all points. A lower score indicates that fewer instruction requirements are satisfied and thus reflects a stronger defense.


\section{Cross-Branch Feature Mapping and Linear CKA}

We use linear centered kernel alignment (CKA) to measure the representational consistency between the ViT and VAE branches. However, the two branches produce different numbers of spatial tokens because their native feature maps have different spatial resolutions. Therefore, before computing CKA, we map both token sequences onto a shared spatial grid to align their token dimensions.

Let the token representation from branch $b\in{v,z}$ be
\begin{equation}
\mathbf{T}_{b}\in\mathbb{R}^{N_b\times D},
\end{equation}
where $v$ and $z$ denote the ViT and VAE branches, respectively. Here, $N_b=H_bW_b$ is the number of spatial tokens, $(H_b,W_b)$ is the native token-grid resolution, and $D$ is the feature dimension.

For each branch, we first reshape the token sequence back into its original two-dimensional spatial layout:
\begin{equation}
\mathbf{T_b}
\in\mathbb{R}^{H_bW_b\times D}
\rightarrow
\mathbf{F}_{b}
\in\mathbb{R}^{1\times D\times H_b\times W_b}.
\end{equation}
Specifically, the token sequence is reshaped into
$H_b\times W_b\times D$, after which the feature dimension is moved to the channel axis. We then resize the spatial feature map to a common resolution
$(H_c,W_c)$ using bilinear interpolation:
\begin{equation}
\widetilde{\mathbf{F}}_{b}=\operatorname{BilinearInterp}
\left(
\mathbf{F}_{b};
H_c,W_c
\right),
\end{equation}
where
$\widetilde{\mathbf{F}}_{b}
\in\mathbb{R}^{1\times D\times H_c\times W_c}$.

Finally, the resized feature map is converted back into a token sequence:
\begin{equation}
\widetilde{\mathbf{F}}_{b}
\in\mathbb{R}^{1\times D\times H_c\times W_c}
\rightarrow
\widetilde{\mathbf{T}}_{b}
\in\mathbb{R}^{H_cW_c\times D}.
\end{equation}

In our implementation, the common spatial resolution is set to
$16\times16$. Consequently, both branches are mapped to 256 spatial tokens:
\begin{equation}
\widetilde{\mathbf{T}}_{v}
\in\mathbb{R}^{256\times D},
\qquad
\widetilde{\mathbf{T}}_{z}
\in\mathbb{R}^{256\times D}.
\end{equation}
This mapping operation aligns only the token dimension and preserves the original feature dimension of each branch. Both token sequences have the same feature dimension $D$.

After token alignment, we center each representation along the token dimension:
\begin{equation}
\mathbf{X}
=
\widetilde{\mathbf{T}}_{v}
-
\operatorname{Mean}
\left(
\widetilde{\mathbf{T}}_{v},
\operatorname{dim}=0
\right),
\end{equation}
\begin{equation}
\mathbf{Y}
=
\widetilde{\mathbf{T}}_{z}
-
\operatorname{Mean}
\left(
\widetilde{\mathbf{T}}_{z},
\operatorname{dim}=0
\right).
\end{equation}

We construct the corresponding linear Gram matrices as
\begin{equation}
\mathbf{K}=\mathbf{X}\mathbf{X}^{\top},
\qquad
\mathbf{L}=\mathbf{Y}\mathbf{Y}^{\top},
\end{equation}
where
$\mathbf{K},\mathbf{L}\in\mathbb{R}^{256\times256}$.
Each Gram-matrix entry represents the similarity between a pair of aligned spatial tokens within the corresponding branch.

The Gram matrices are further centered using
\begin{equation}
\mathbf{K}_{c}=\mathbf{H}\mathbf{K}\mathbf{H},
\qquad
\mathbf{L}_{c}=\mathbf{H}\mathbf{L}\mathbf{H},
\end{equation}
where
\begin{equation}
\mathbf{H}
=
\mathbf{I}
-
\frac{1}{N}\mathbf{1}\mathbf{1}^{\top},
\qquad
N=H_cW_c.
\end{equation}

The linear CKA similarity is then computed as
\begin{equation}
\operatorname{CKA}(\mathbf{X},\mathbf{Y})
=
\frac{
\langle\mathbf{K}_{c},\mathbf{L}_{c}\rangle_{F}
}{
|\mathbf{K}_{c}|_{F}
|\mathbf{L}_{c}|_{F}
+\epsilon
},
\end{equation}
where $\langle\cdot,\cdot\rangle_F$ denotes the Frobenius inner product, $|\cdot|_F$ denotes the Frobenius norm, and $\epsilon$ is a small constant for numerical stability.

Through this procedure, the ViT and VAE representations are compared according to their internal spatial-relation structures after being aligned to the same number of tokens. A higher CKA value indicates stronger structural consistency between the two branches, whereas a lower value indicates that their cross-branch representational consistency has been disrupted.

\section{Implementation Details}

\subsection{Details of Optimization and Editing}
\paragraph{Adversarial Example Generation and Image Editing.}
All experiments were conducted on a single NVIDIA A100 GPU with 80GB of memory. We adopted BAGEL-7B-MoT as the surrogate model. BAGEL employs a Mixture-of-Transformers (MoT) architecture with two separate visual encoding pathways. Specifically, it uses a SigLIP2 Vision Transformer (ViT) to extract semantic-level visual representations and a pretrained VAE from FLUX to encode pixel-level information into the latent space. We mainly conducted experiments on the VGGFace2 and CelebA-HQ datasets. All input images were resized to a unified resolution of $512 \times 512 \times 3$. During adversarial example generation, the maximum perturbation budget was set to $\epsilon = 8/255$. For each baseline method, we followed the default hyperparameter settings reported in its original paper. The textual prompt provided to BAGEL was set to an empty string during the generation of adversarial example.

To evaluate the protection performance of the adversarial examples, we used the image-editing mode of BAGEL and performed image editing under four different instructions. For our method, the loss-weighting hyperparameters $\alpha$, $\beta$, and $\gamma$ were set to $10$, $0.2$, and $1$, respectively. When computing linear centered kernel alignment (CKA), the spatial resolution of the interpolated feature maps was set to $16 \times 16$. All seeds are set to 2026.

\subsection{Evaluation Metrics}

We evaluate the protection effectiveness of adversarial examples from three complementary perspectives: the quality of the edited images, compliance with the editing instructions, and facial identity similarity.

\paragraph{Fréchet Inception Distance (FID).}
Fréchet Inception Distance (FID) measures the distributional discrepancy between two sets of images in the feature space of a pretrained Inception network. In our experiments, we compute FID between the images edited from clean inputs and those edited from their adversarial counterparts. A higher FID score indicates a larger distributional discrepancy between the two sets of edited images, suggesting that the adversarial perturbations more strongly interfere with the image-editing process and therefore provide better protection.

\paragraph{Blind/Referenceless Image Spatial Quality Evaluator (BRISQUE).}
BRISQUE is a no-reference image-quality assessment metric that evaluates perceptual image quality without requiring a corresponding reference image. It estimates image degradation based on deviations from natural-scene statistics. A higher BRISQUE score indicates poorer perceptual quality of the edited image. In our setting, a higher score suggests that the adversarial example more effectively disrupts the image-generation process.

\paragraph{Identity Similarity Metric (ISM).}
To determine whether the identity of the original subject is preserved in the edited image, we first use RetinaFace to detect and localize the facial regions. We then extract identity-specific representations from the detected faces and compute the similarity between the original clean image and the corresponding edited image. A lower ISM score indicates that the face in the edited image is less similar to the original identity, demonstrating stronger protection against unauthorized identity-preserving image editing.


\section{Transferability}
We further evaluate adversarial examples optimized on BAGEL against InternVL-U without target-model re-optimization. Under both $\epsilon=8/255$ and $\epsilon=16/255$, CCS retains a measurable protection effect, demonstrating that the learned perturbations capture certain model-shared vulnerabilities and possess cross-model transferability. Nevertheless, the protection performance decreases compared with the grey-box results on BAGEL. This degradation is expected because BAGEL and InternVL-U differ in their visual encoders, tokenization and feature-alignment mechanisms, multimodal fusion architectures, and generation dynamics. Consequently, perturbations optimized to disrupt BAGEL’s ViT–VAE representations and their cross-branch consistency may not induce equally strong conflicts in InternVL-U. Despite this architectural gap, the observed protection effect without access to the target model provides evidence that CCS is not entirely tied to BAGEL-specific features.
\begin{table}[t]
    \centering
    \caption{Transferability of CCS from BAGEL to InternVL-U under different perturbation budgets. Lower is better.}
    \label{tab:transferability}
    
    \setlength{\tabcolsep}{4pt}
    \renewcommand{\arraystretch}{1.10}
    
    \resizebox{\columnwidth}{!}{
    \begin{tabular}{lcccc}
        \toprule
        \multirow{2}{*}{\textbf{Model}}
        & \multicolumn{2}{c}{$\epsilon=8/255$}
        & \multicolumn{2}{c}{$\epsilon=16/255$} \\
        \cmidrule(lr){2-3}
        \cmidrule(lr){4-5}
        & \textbf{ISM} ($\downarrow$)
        & \textbf{Und.} ($\downarrow$)
        & \textbf{ISM} ($\downarrow$)
        & \textbf{Und.} ($\downarrow$) \\
        \midrule
        BAGEL
        & 0.00 & 0.00
        & 0.00 & 0.00 \\
        
        InternVL-U
        & 0.23 & 81.23
        & 0.18 & 79.0 \\
        \bottomrule
    \end{tabular}
    }
\end{table}
\section{Results on CelebaHQ dataset}

We follow the settings in the main text on CelebaHQ. Results are in Table~\ref{tab:method_cele}, Table~\ref{tab:method_cele_p2}, Table~\ref{tab:method_cele_p3} and Table~\ref{tab:method_cele_p4}. This demonstrates that the effectiveness of our method is not limited to specific datasets. We follow the experimental settings in the main text and further evaluate CCS on CelebA-HQ. As reported in Tables 3–6, CCS consistently achieves substantially higher FID and markedly lower Und-Score across all four editing prompts. In particular, its Und-Score remains between 1.00 and 2.50, whereas the baseline methods obtain scores ranging from 54.17 to 99.17, indicating that CCS effectively prevents the model from following diverse editing instructions. CCS also generally produces stronger output-quality degradation, as reflected by its competitive BRISQUE results. These consistent improvements across prompts demonstrate that CCS remains effective on CelebA-HQ and that its protection performance is not limited to a particular facial dataset or editing instruction.

\begin{table*}[t]
    \centering

    \begin{minipage}[t]{0.49\textwidth}
        \centering
        \setlength{\tabcolsep}{3pt}
        \resizebox{\linewidth}{!}{
        \begin{tabular}{lcccc}
            \toprule
            \multirow{2}{*}{\textbf{Methods}} & \multicolumn{4}{c}{\textbf{Prompt~1}} \\ 
            \cmidrule(lr){2-5}
            & \textbf{ISM($\downarrow$)} & \textbf{FID($\uparrow$)} & \textbf{BRISQUE($\uparrow$)} & \textbf{Und-Score($\downarrow$)} \\ 
            \midrule
            AdvDM       & 0.37 & 174.31 & 34.22 & 96.00 \\ 
            
            Mist        & 0.22 & 270.18 & 42.59 & 65.50 \\ 
            PID         & 0.13 & 213.50 & \textbf{47.50} & 67.50 \\ 

            \textbf{CCS (ours)} & \textbf{0.00} & \textbf{461.09} & 42.81 & \textbf{1.00} \\ 
            \bottomrule
        \end{tabular}
        }
        \captionof{table}{Performance comparison of different methods on prompt 1: ``The person wore a smile, holding a book in hands-with `BAGEL' on its cover''.}
        \label{tab:method_cele}
    \end{minipage}
    \hfill
    \begin{minipage}[t]{0.49\textwidth}
        \centering
        \setlength{\tabcolsep}{3pt}
        \resizebox{\linewidth}{!}{
        \begin{tabular}{lcccc}
            \toprule
            \multirow{2}{*}{\textbf{Methods}} & \multicolumn{4}{c}{\textbf{Prompt 2}} \\ 
            \cmidrule(lr){2-5}
            & \textbf{ISM($\downarrow$)} & \textbf{FID($\uparrow$)} & \textbf{BRISQUE($\uparrow$)} & \textbf{Und-Score($\downarrow$)} \\ 
            \midrule
            AdvDM       & 0.13 & 119.74 & 21.19 & 95.00 \\ 
            
            Mist        & 0.06 & 226.52 & 30.68 & 66.87 \\ 
            PID         & 0.06 & 210.09 & 30.48 & 61.25 \\ 

            \textbf{CCS (ours)} & \textbf{0.00}  & \textbf{467.35} & \textbf{40.16} & \textbf{2.50} \\ 
            \bottomrule
        \end{tabular}
        }
        \captionof{table}{Performance comparison of different methods on prompt 2: ``The person is holding an umbrella on a rainy street''.}
        \label{tab:method_cele_p2}
    \end{minipage}

    \vspace{0.8em}

    \begin{minipage}[t]{0.49\textwidth}
        \centering
        \setlength{\tabcolsep}{3pt}
        \resizebox{\linewidth}{!}{
        \begin{tabular}{lcccc}
            \toprule
            \multirow{2}{*}{\textbf{Methods}} & \multicolumn{4}{c}{\textbf{Prompt~3}} \\ 
            \cmidrule(lr){2-5}
            & \textbf{ISM($\downarrow$)} & \textbf{FID($\uparrow$)} & \textbf{BRISQUE($\uparrow$)} & \textbf{Und-Score($\downarrow$)} \\ 
            \midrule
            AdvDM       & 0.13 & 132.19 & 16.96 & 99.17 \\ 
            
            Mist        & 0.26 & 213.29 & 34.69 & 65.83 \\ 
            PID         & 0.22 & 199.34 & 29.86 & 65.00 \\ 

            \textbf{CCS (ours)} & \textbf{0.00} & \textbf{421.92} & \textbf{44.96} & \textbf{1.67} \\ 
            \bottomrule
        \end{tabular}
        }
        \captionof{table}{Performance comparison of different methods on Prompt~3: ``The person is holding a bouquet of flowers''.}
        \label{tab:method_cele_p3}
    \end{minipage}
    \hfill
    \begin{minipage}[t]{0.49\textwidth}
        \centering
        \setlength{\tabcolsep}{3pt}
        \resizebox{\linewidth}{!}{
        \begin{tabular}{lcccc}
            \toprule
            \multirow{2}{*}{\textbf{Methods}} & \multicolumn{4}{c}{\textbf{Prompt~4}} \\ 
            \cmidrule(lr){2-5}
            & \textbf{ISM($\downarrow$)} & \textbf{FID($\uparrow$)} & \textbf{BRISQUE($\uparrow$)} & \textbf{Und-Score($\downarrow$)} \\ 
            \midrule
            AdvDM       & 0.03 & 94.73 & 16.51 & 76.67 \\ 
            
            Mist        & 0.05 & 164.10 & 29.86 & 55.83 \\ 
            PID         & 0.05 & 139.11 & 36.45 & 54.17 \\ 

            \textbf{CCS (ours)} & \textbf{0.00}  & \textbf{426.55} & \textbf{40.82} & \textbf{2.50} \\ 
            \bottomrule
        \end{tabular}
        }
        \captionof{table}{Performance comparison of different methods on Prompt~4: ``The person is standing by the Eiffel Tower''.}
    \label{tab:method_cele_p4}
    \end{minipage}

\end{table*}

\section{Hyper-parameter Analysis}
\subsection{Interpolation Size form Mapping Function}
\begin{figure}[t]
    \includegraphics[width=\columnwidth]{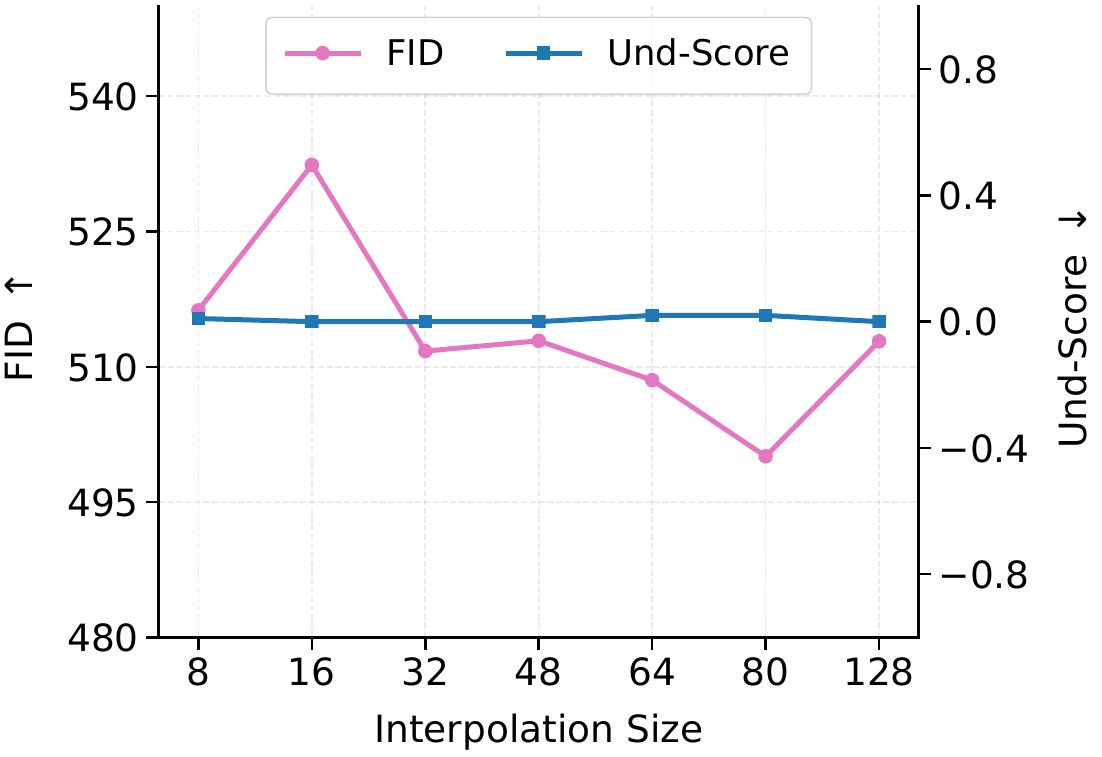}
    \caption{Effect of the interpolation size in the mapping function. Our method maintains consistently high FID and near-zero Und-Score across different sizes, demonstrating low sensitivity to this hyperparameter.}
    \label{fig:mapping_size}
\end{figure}

As shown in Figure~\ref{fig:mapping_size}, our method is largely
insensitive to the interpolation size used in the mapping function. Across a wide range of sizes, the FID remains consistently high while the Und-Score (range: 0-100) stays close to zero, demonstrating stable protection performance without careful hyperparameter tuning. We therefore use $16\times16$ as the default setting for a favorable balance between effectiveness and computational efficiency.
\subsection{Weight Parameter on Branch-wise Design}
\begin{table}[t]
    \centering
    \caption{Ablation study on the weights of the VAE and ViT branch objectives.
    Lower ISM and Und-Score indicate stronger protection performance.}
    \label{tab:branch_weight_ablation}
    \setlength{\tabcolsep}{10pt}
    \renewcommand{\arraystretch}{1.15}
    \begin{tabular}{lccc}
        \toprule
        \textbf{Branch} & \textbf{Weight} 
        & \textbf{ISM} ($\downarrow$) 
        & \textbf{Und-Score} ($\downarrow$) \\
        \midrule

        \multirow{4}{*}{VAE}
        & 0    & 0.49 & 76.00 \\
        & 2.5  & 0.01 & 6.00  \\
        & 5    & 0.00 & 2.75 \\
        & 15   & 0.00 & 0.00 \\
        \midrule

        \multirow{4}{*}{ViT}
        & 0    & 0.00 & 1.25  \\
        & 0.5  & 0.00 & 3.25  \\
        & 1    & 0.01 & 6.25  \\
        & 5    & 0.13 & 29.50 \\
        
        \bottomrule
    \end{tabular}
\end{table}
We conduct a hyperparameter analysis of the weights assigned to the ViT and VAE objectives. We find that increasing the VAE weight improves the overall protection performance but reduces the disruption to the understanding branch.

\section{More Cases}
\begin{figure*}[t]
    \centering
    \includegraphics[width=\linewidth]{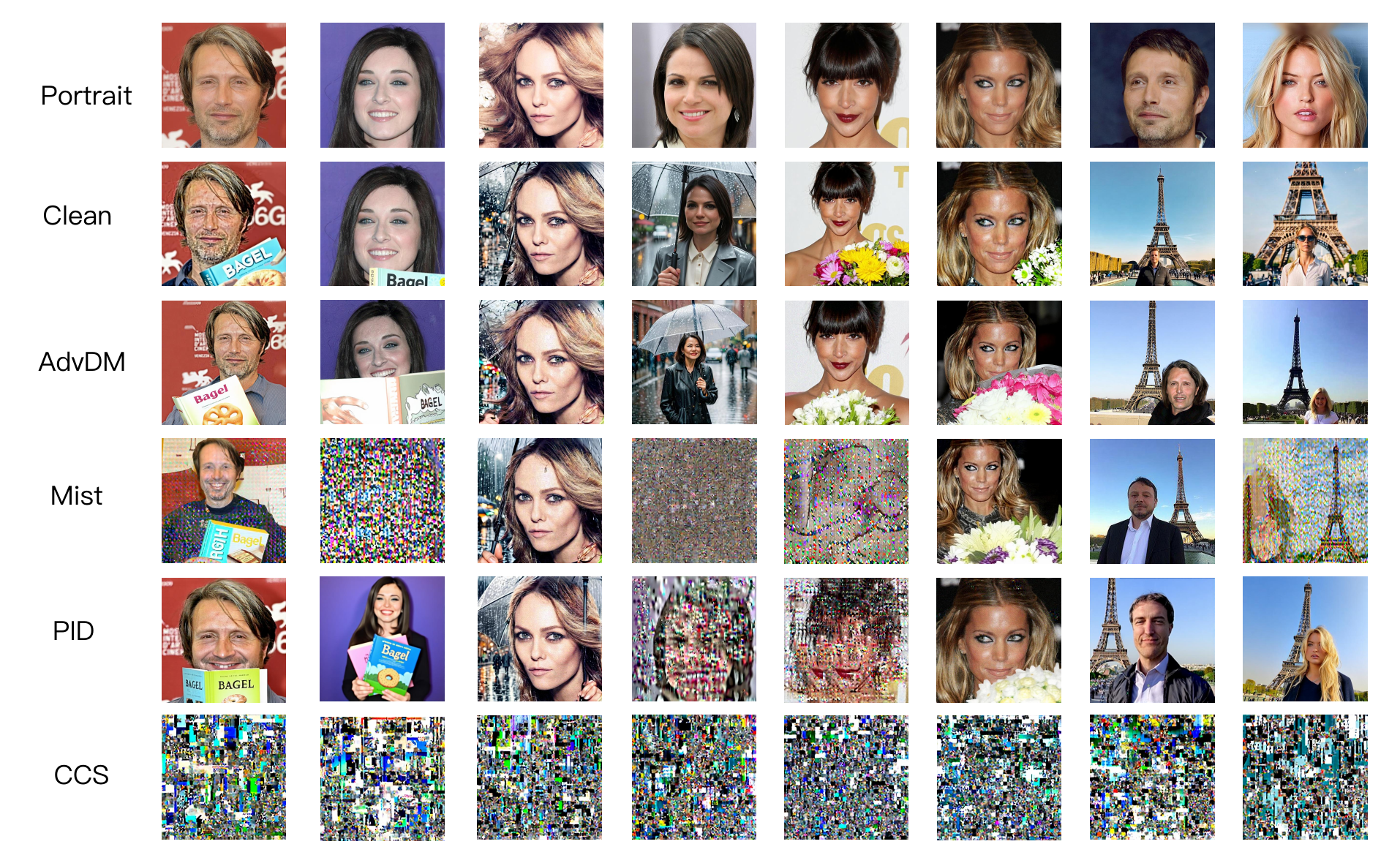}
    \caption{Qualitative results of our method v.s. baselines. CCS demonstrates substantially stronger protection. }
    \label{fig:celebahq_results}
\end{figure*}

As shown in Fig.~\ref{fig:celebahq_results}, clean images and existing protection methods generally allow the model to generate realistic edited portraits that preserve recognizable facial identities and follow the given instructions, such as holding a book, carrying an umbrella, holding flowers, or appearing in front of the Eiffel Tower. Although Mist and PID occasionally introduce visible artifacts, their protection effects are inconsistent across identities and prompts, and some outputs remain successfully edited. In contrast, CCS consistently disrupts the generation process across all tested examples, producing severely distorted outputs in which neither coherent facial identity nor the requested editing semantics can be recovered. These qualitative observations are consistent with the quantitative results and further demonstrate the strong and stable protection performance of CCS on CelebA-HQ.

\section{Extension to Non-Facial Images}
\begin{figure*}[t]
    \centering
    \includegraphics[width=\linewidth]{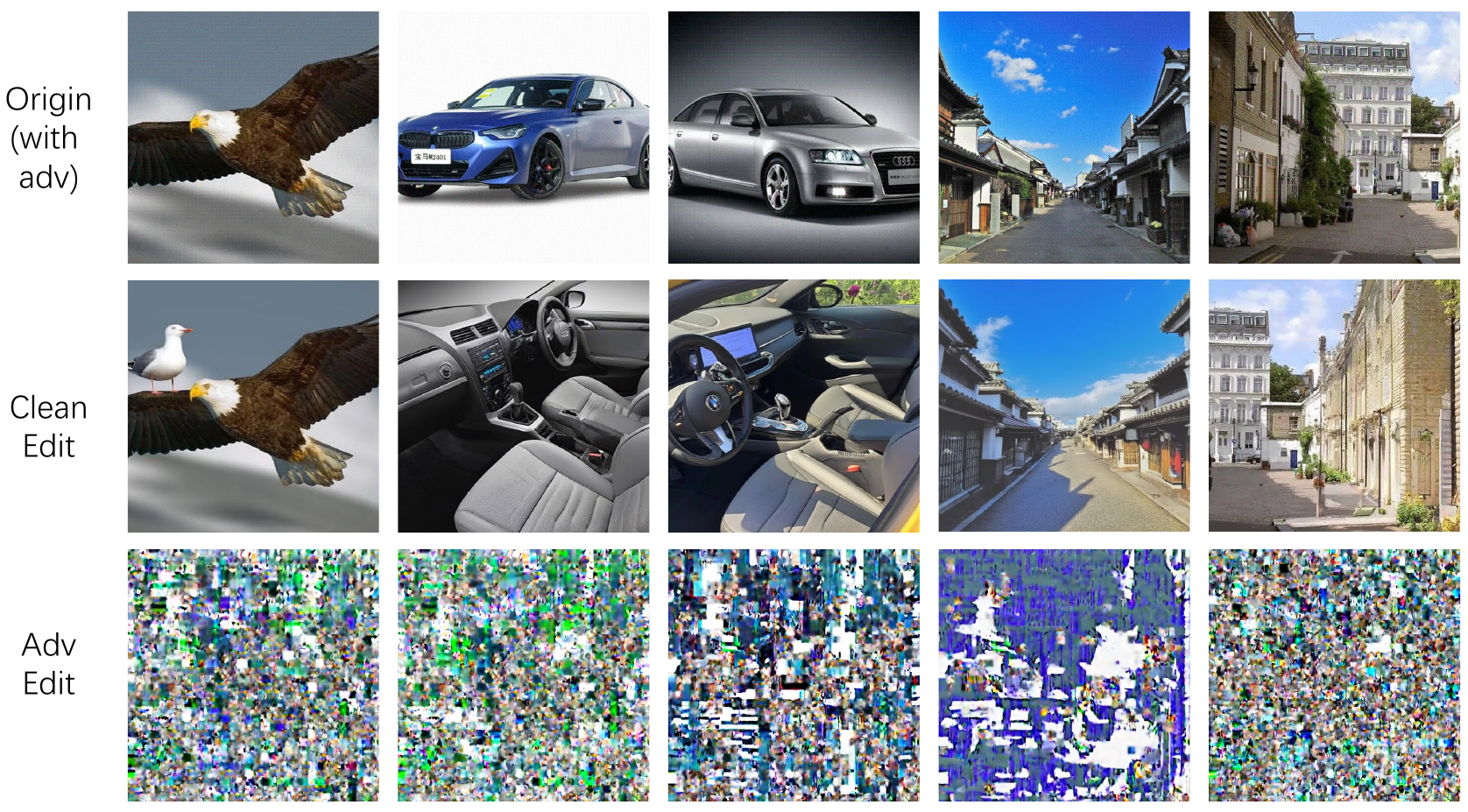}
    \caption{Qualitative results of our method on non-facial editing tasks. CCS also demonstrates stronger protection. }
    \label{fig:non_facial_results}
\end{figure*}

We further collect several non-facial datasets to evaluate the generalizability of our method in Figure~\ref{fig:non_facial_results}. Our adversarial perturbations are not limited to disrupting facial image editing, but also effectively interfere with the editing of non-facial images. For example, in scenarios involving BAGEL's world-understanding and general-purpose image-editing capabilities, our adversarial examples likewise cause the model to fail to perform the requested edits. For vehicle images, we instruct the UMM to generate the corresponding interior scenes. For street-view images, we evaluate the UMM’s physical understanding ability using instructions such as “turn right.” The UMM can successfully generate images consistent with these instructions. However, after applying CCS perturbations to the input images, the model can no longer extract reliable visual features, leading to clear editing failures.




\end{document}